\documentclass[lettersize,journal]{IEEEtran}
\usepackage{amsmath,amsfonts}
\usepackage{algorithmic}
\usepackage{algorithm}
\usepackage{array}
\usepackage[caption=false,font=normalsize,labelfont=sf,textfont=sf]{subfig}
\usepackage{textcomp}
\usepackage{stfloats}
\usepackage{url}
\usepackage{verbatim}
\usepackage{graphicx}
\usepackage{cite}
\usepackage{acro}
\usepackage{hyperref}

\newcommand\numberthis{\addtocounter{equation}{1}\tag{\theequation}}

\DeclareAcronym{UAV}{
  short=UAV,
  long=Unmanned Aerial Vehicle
}
\DeclareAcronym{RL}{
  short=RL,
  long=Reinforcement Learning
}
\DeclareAcronym{ROS}{
  short=ROS,
  long=Robot Operating System
}
\DeclareAcronym{SITL}{
  short=SITL,
  long=Software-In-The-Loop
}
\DeclareAcronym{HITL}{
  short=HITL,
  long=Hardware-In-The-Loop
}
\DeclareAcronym{URDF}{
  short=URDF,
  long=Unified Robot Description Format
}
\DeclareAcronym{DAE}{
  short=DAE,
  long=Digital Asset Exchange
}
\DeclareAcronym{YAML}{
  short=YAML,
  long=YAML Ain't Markup Language
}
\DeclareAcronym{ENU}{
  short=ENU,
  long=East-North-Up
}
\DeclareAcronym{PID}{
  short=PID,
  long=Proportional Integral Derivative
}
\DeclareAcronym{CI}{
  short=CI,
  long=Continuous Integration
}
\DeclareAcronym{RNG}{
  short=RNG,
  long=Random Number Generator
}
\DeclareAcronym{PyPI}{
  short=PyPI,
  long=The Python Package Index
}
\DeclareAcronym{CFD}{
  short=CFD,
  long=Computational Fluid Dynamics
}
\DeclareAcronym{AoA}{
  short=AoA,
  long=Angle of Attack
}
\DeclareAcronym{MDP}{
  short=MDP,
  long=Markov Decision Process
}
\DeclareAcronym{SAC}{
  short=SAC,
  long=Soft Actor Critic
}
\DeclareAcronym{CCGE}{
  short=CCGE,
  long=Critic Confidence Guided Exploration
}
\DeclareAcronym{AWAC}{
  short=AWAC,
  long=Advantage Weighted Actor Critic
}

\begin{document}

\title{PyFlyt---UAV Simulation Environments for Reinforcement Learning Research}

\author{Jun Jet Tai, Jim Wong, Mauro S. Innocente, Nadjim Horri, James Brusey~\IEEEmembership{Member, IEEE}, Swee~King~Phang~\IEEEmembership{Senior~Member,~IEEE}
\thanks{J. J. Tai, J. Wong, M. Innocente and J. Brusey are with Coventry University, Priory St, Coventry CV1 5FB, United Kingdom (emails: taij@uni.coventry.ac.uk, wongjim8@gmail.com, mauro.s.innocente@coventry.ac.uk, james.brusey@coventry.ac.uk)}
\thanks{N. Horri is with the School of Engineering, University of Leicester, University Road, Leicester LE1 7RH, United Kingdom (email: nmhorri@hotmail.com)}
\thanks{S. K. Phang is with the School of Engineering, Taylor's University, Subang Jaya 47500, Malaysia (email: sweeking.phang@taylors.edu.my)}}

\markboth{Journal of \LaTeX\ Class Files,~Vol.~14, No.~8, August~2021}%
{Tai \MakeLowercase{\textit{et al.}}: PyFlyt - UAV Simulation Environments for RL Research}

\IEEEpubid{0000--0000/00\$00.00~\copyright~2021 IEEE}

\maketitle

\begin{abstract}
Unmanned aerial vehicles (UAVs) have numerous applications, but their efficient and optimal flight can be a challenge.
\ac{RL} has emerged as a promising approach to address this challenge, yet there is no standardized library for testing and benchmarking RL algorithms on UAVs.
In this paper, we introduce PyFlyt, a platform built on the Bullet physics engine with native Gymnasium API support.
PyFlyt provides modular implementations of simple components, such as motors and lifting surfaces, allowing for the implementation of UAVs of arbitrary configurations.
Additionally, PyFlyt includes various task definitions and multiple reward function settings for each vehicle type.
We demonstrate the effectiveness of PyFlyt by training various RL agents for two UAV models: quadrotor and fixed-wing.
Our findings highlight the effectiveness of RL in UAV control and planning, and further show that it is possible to train agents in sparse reward settings for UAVs.
PyFlyt fills a gap in existing literature by providing a flexible and standardised platform for testing RL algorithms on UAVs.
We believe that this will inspire more standardised research in this direction.

\end{abstract}


\section{Introduction}

\IEEEPARstart{R}{einforcement} Learning has rapidly advanced in the past decade, achieving various impressive feats \cite{silver2017mastering, silver2018general, berner2019dota, vinyals2019grandmaster, degrave2022magnetic}.
Fundamentally, it's allure boils down to its theoretical simplicity and generalisability, with growth fueled by open and accessible frameworks and software.
Most notably, various environment API libraries \cite{brockman2016openai, dm_env2019} and learning frameworks \cite{raffin2019stable, liang2018rllib, huang2021cleanrl} have made standardising \ac{RL} across a large range of tasks possible.

\begin{figure}
    \centering
    \includegraphics[width=0.4\textwidth]{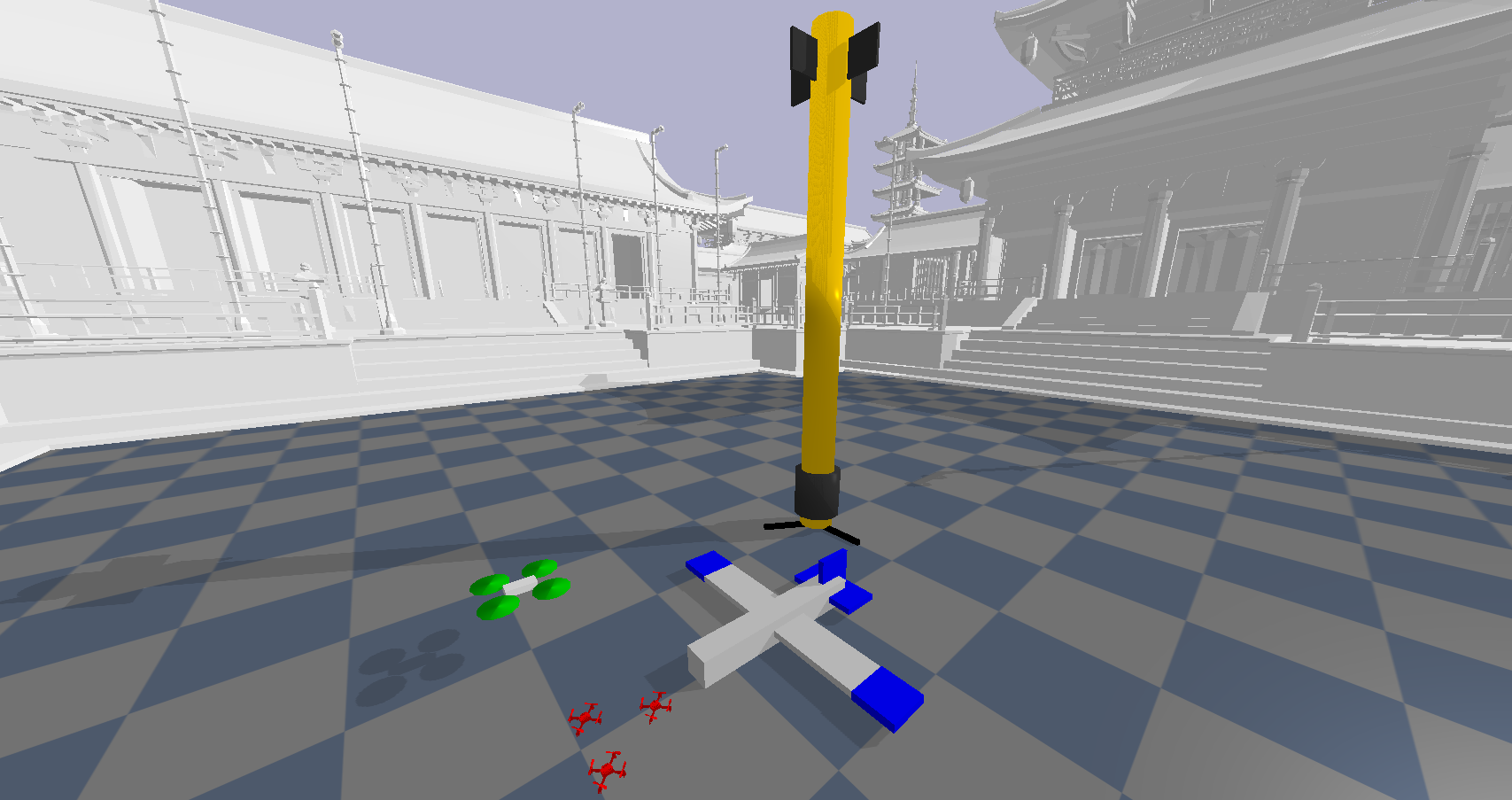}
    \caption{The PyFlyt simulator, showing some prebuilt \acp{UAV} - one Rocket, one Fixedwing, one generic QuadX, and a cluster of three Crazyflie QuadXs.}
    \label{pyflyt_cover}
\end{figure}

\ac{RL} draws inspiration from control theory, evidenced by several \ac{RL} tasks mimicking those in control theory.
This also holds true for \ac{UAV} flight, which has long held the interest of the controls engineering community.
The classical approach to optimising \ac{UAV} control involves testing in simulation.
Popular open-source \ac{UAV} simulation tools include Gazebo \cite{koenig2004design}, FlightGear \cite{perry2004flightgear}, X-Plane \cite{garcia2010multi}, AirSim \cite{shah2018airsim} and UE4Sim \cite{mueller2017ue4sim}.
However, their limited portability severely restricts their use in machine learning workflows, which usually run on highly parallel remote computing clusters.
Moreover, with the exception of fully fledged robotics simulators like Gazebo, simulation is limited to one or two variants of \acp{UAV}, reducing their adaptability for \ac{RL} environments using uncommon aircraft architectures like the quadplane \cite{mathur2021design}. Additionally, simulators like UE4Sim lack a physics model, rendering it impossible to modify physical \ac{UAV} attributes such as motor thrust profile or moments of inertia.

In this work, we present an extensible and open source \ac{UAV} design and simulation library named \textit{PyFlyt} (pronounced pa\textsc{I}-fla\textsc{I}t) that fills the void between complex \ac{UAV} simulators and user-friendly machine learning libraries described in Section \ref{related_works}.
More formally, PyFlyt offers the following key features:
\IEEEpubidadjcol
\begin{itemize}
    \item \textit{Ease of Use}: automatic handling of various physics, control and agent loop rates, collisions, multiple \acp{UAV} in the same environment, plus implementation of various classical control algorithms for low level flight control and extensible dynamics via the Bullet Physics Engine (Section \ref{overall_architecture}).
    \item \textit{Configurability}: flexible construction of \acp{UAV} using modular implementations of basic components such as motors, boosters, lifting surfaces, gimbals and cameras (Section \ref{uav_component_descriptions}).
    \item \textit{Standardisation}: pre-implemented designs for various \acp{UAV} (Section \ref{uav_model_description}) along with corresponding environments compatible with the Gymnasium API (Section \ref{gymnasium_environments}), tested extensively with various \ac{RL} algorithms (Section \ref{reinforcement_learning_results}).
    \item \textit{Quality of Life Improvements}: deterministic simulation depending on initialisation seed, \ac{CI} testing on the master repository and a packaging and release cycle to \ac{PyPI}.
\end{itemize}

\section{Related Works} \label{related_works}

\ac{RL} environments and \ac{UAV} simulators are not a new; many already exist with various functionality and capabilities.
In this section, we briefly cover the landscape of \ac{RL} environment libraries and \ac{UAV} simulation tools, including those catered for \ac{RL}.
At the end, we draw similarities between these tools and PyFlyt, and highlight the gap we aim to fill.

\subsection{Reinforcement Learning Environments}

OpenAI's release of the Gym library in 2016 \cite{brockman2016openai} standardized benchmarking and interfacing for \ac{RL}.
Subsequently, various \ac{RL} environment libraries built on the Gym API have emerged, including those based on video games \cite{kempka2016vizdoom, milani2020retrospective} or classic robotics problems \cite{koch2019reinforcement, lopez2019gym}
The original OpenAI Gym paper has been cited over 5000 times, and hundreds of well-documented and widely used environment libraries have since emerged, despite the existence of other APIs \cite{dm_env2019}.

Recently, volunteer contributors to Gym have moved to Gymnasium, an open-source fork of Gym that will receive all further maintenance and development moving forward \cite{farama_announcing_2022}.
Gymnasium offers more official support for the \texttt{termination} and \texttt{truncation} API\footnote{\url{https://github.com/openai/gym/issues/2510}} and has seen popular libraries like Gymnasium Robotics \cite{gymnasium_robotics2023github}, ViZDoom \cite{kempka2016vizdoom}, Minigrid \cite{minigrid}, MiniWoB++ \cite{shi2017world}, and MicroRTS-Py \cite{huang2021gym} make the transition to its platform.
In addition, a wrapper library by the authors of Gymnasium called Shimmy \cite{shimmy2022github} allows for the conversion of environments in other APIs such as DM-Env and DM-Control into a Gymnasium-compatible one.
The active development of Gymnasium and its constant adoption by other libraries make it an easy choice for PyFlyt to be built around the its API.

\subsection{UAV Simulators}

When selecting a \ac{UAV} simulator for a project, several factors need to be considered, including visual realism, physics requirements like turbulence or downwash, and \ac{HITL} capabilities, among others.
Commercially available simulators like XPlane and FlightGear can simulate a wide range of fixed-wing aircraft with pseudo-accurate world scenery.
These simulators offer plugin systems that enable users to create extension modules, resulting in various research tools, including those that can be used with Matlab/Simulink for developing flight controllers \cite{ribeiro2010uav, bittar2014guidance, figueiredo2012simulation, sorton2005simulated, ying2017visual}.
Some of these tools have also been adapted for OpenAI's Gym API interface \cite{richter2021qplane, chin2020gymfg}.
However, despite their capabilities, these simulators have complicated installation procedures, minimal Linux support, and require graphical user interfaces, making them less suitable for most machine learning workflows.

Multirotor simulators like JMavSim and Microsoft's AirSim \cite{shah2018airsim} are tools designed specifically for multirotor \ac{UAV} simulation.
Both software packages offer \ac{SITL} and \ac{HITL} simulation with the PX4 \cite{meier2015px4} flight control firmware and \ac{ROS}.
AirSim, being the most popular, offers photorealistic rendering through Unreal Engine 4 at the expense of elevated computational requirements.
However, creating custom multirotor \ac{UAV} models for these simulators is challenging, making them less ideal when designing environments for \ac{RL} algorithms revolving around \ac{UAV}'s with different parameters, such as thrust profile or moment of inertia.

The most closely related work to ours is gym-pybullet-drones \cite{panerati2021learning} and PyFly \cite{bohn2019deep} (PyFlyt's name was chosen before we knew of PyFly).
Like PyFlyt, Gym-pybullet-drones is a Bullet Physics Engine based simulator, but it uses the older OpenAI Gym API and supports simulation for only the Crazyflie 2.0 \ac{UAV} \cite{giernacki2017crazyflie}.
PyFly, on the other hand, is a Python implementation of a 6 degree-of-freedom aerodynamic model for fixed-wing aircraft.
Like gym-pybullet-drones, PyFly uses the older OpenAI Gym API, but lacks rendering capabilities and only  has \ac{RL} environments for attitude control.
In comparison, PyFlyt allows for the construction of \textit{both} multirotor and fixedwing \acp{UAV}, in arbitrary configurations such as the trirotor, hexarotor, tube-and-wing, flying-wing or biplane.

Overall, existing simulators provide various capabilities, but they also have limitations that make them less suitable for certain applications.
PyFlyt aims to address these limitations by providing a more flexible and user-friendly platform for running \ac{RL} algorithms on a range of configurable \ac{UAV} platforms, all within a single library \footnote{Available for install with a single command from: \url{https://pypi.org/project/PyFlyt/}}.

\section{Design of PyFlyt} \label{pyflyt}

In this section, we break down the design methodology of PyFlyt by first describing its main goals in Section \ref{main_goals} before going into the overall architecture in Section \ref{overall_architecture}.
The detail the modular \ac{UAV} components in Section \ref{uav_component_descriptions} and how they form the basic \ac{UAV} platforms in Section \ref{uav_model_description}.
Finally, we describe the default Gymnasium compatible environments in Section \ref{gymnasium_environments} and describe potential extensions, possibilities, and improvements in Section \ref{possible_extensions}.

\subsection{Main Goals} \label{main_goals}

PyFlyt is a library primarily designed for researchers to develop and test complex \ac{RL} algorithms for \acp{UAV}, but the development of classical control algorithms is also well within scope.
It treats single agent simulations as multiagent simulations with a single agent.
This allows easy extension to the multiagent setting.
PyFlyt also automatically tracks collisions between different entities, allowing the testing and evaluation of collision avoidance algorithms either with obstacles or with other \acp{UAV}.

Furthermore, PyFlyt supports the construction of of arbitrary \ac{UAV} configurations by combining various basic components.
This includes the addition of onboard cameras --- gimballed or fixed --- with RGB, depth maps, and segmentation maps, allowing for the development of vision-based \ac{RL} algorithms.

Moreover, there is built-in support for developing generic onboard flight control.
This simplifies the process of developing and testing \ac{RL} algorithms by allowing researchers to map \ac{RL} agent actions to setpoints in the flight controller, rather than directly controlling raw actuator outputs.
In addition to that, the library supports running \ac{RL}, control, and physics at different frequencies.
This leads to more realistic simulation, as the \ac{RL} loop is unlikely to always be running at the fastest looprate.
This capability is useful when doing research on a more macro level, such as when studying swarming behaviours \cite{innocente2019self}.

Finally, the creation of Gymnasium compatible environments using PyFlyt is done as an auxiliary module instead of being integrated into the core architecture of PyFlyt.
This is in contrast to the implementation in gym-pybullet drones, allowing researchers to utilize PyFlyt to construct arbitrary environments using any desired \ac{RL} API.
Several interesting and well tested environments have also been implemented for various vehicles.
Their goal is to serve as a good baseline for upcoming research, as well as to serve as templates for users aiming to create their own environments.

\subsection{Overall Architecture} \label{overall_architecture}

PyFlyt utilizes the Bullet Physics Engine, chosen for its open-source nature, fast C++ backend, flexible rendering options, and time-discrete steppable physics, Python bindings, and support for various robot file definitions.
It was designed with extensibility in mind, allowing users to customize the simulation environment without relying on inheritance-based approaches \cite{alam2012designing}.

\begin{figure}
    \centering
    \includegraphics[width=0.4\textwidth]{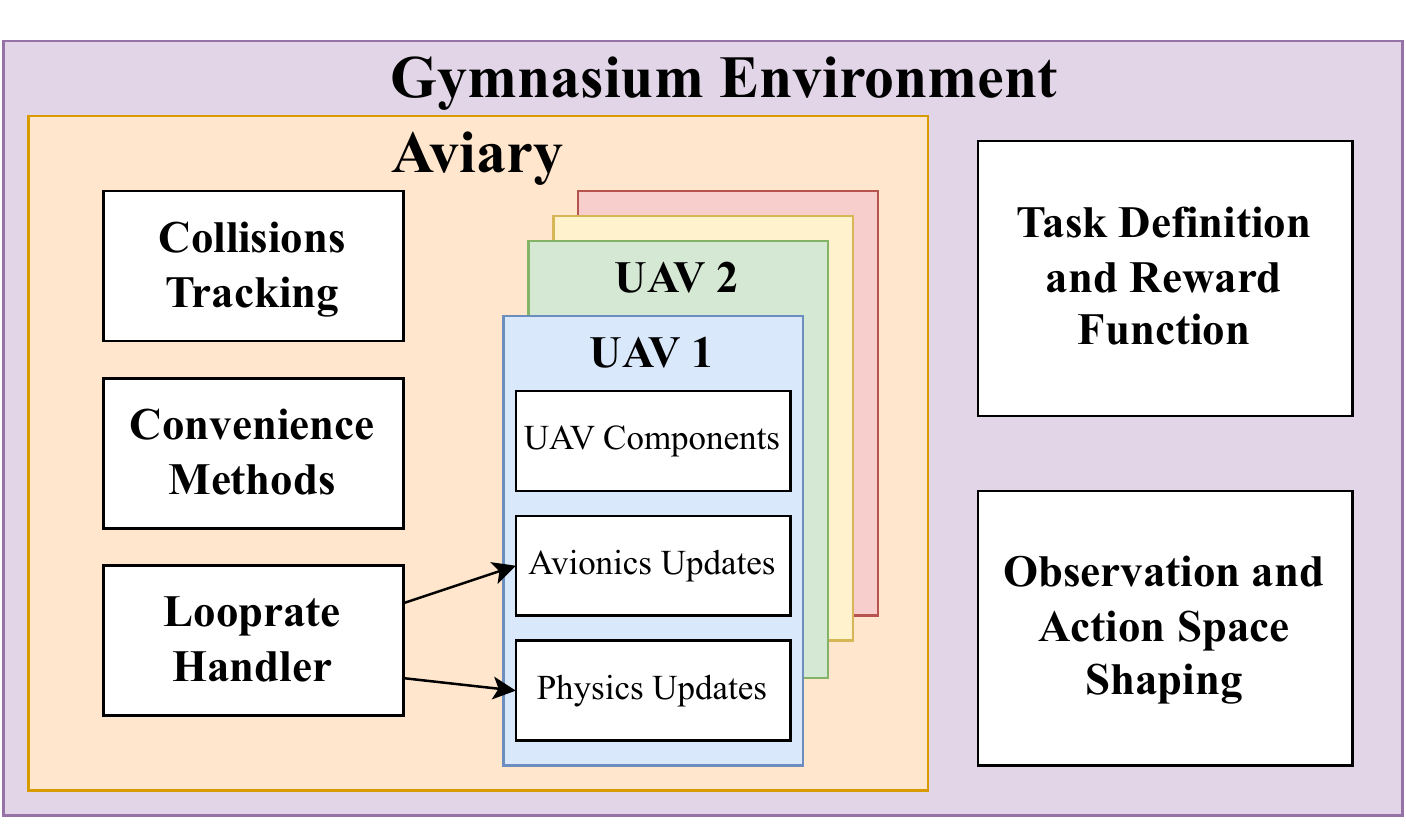}
    \caption{Top level overview of the PyFlyt architecture, showing the composition of classes that make up core \ac{UAV} flight engine and how it integrates into a Gymnasium environment. Coloured boxes describe a class definition, while white boxes describe various functionality of the classes.}
    \label{pyflyt_architecture}
\end{figure}

The library's architecture is composed of various classes, shown in Figure \ref{pyflyt_architecture}.
At the core lies the Aviary, serving as a domain for all \acp{UAV}.
The Aviary handles collision tracking between entities and performs step scheduling according to the various looprates.
It also includes convenience functions for setting setpoints, arming status, and flight modes for all \acp{UAV}.

The Aviary can accommodate any number of \acp{UAV}, each defined by a combination of a \ac{URDF} file and a Python class.
The \ac{URDF} file describes the geometrical and physical properties of the UAV, including its mass, inertia, as well as visual and collision meshes, which can be imported as \ac{DAE} files from standard CAD software.
The Python class associated with each \ac{UAV} inherits from a common Drone Class, which defines only the core functionality shared by all \acp{UAV}, necessary for interfacing with the Aviary.
Each \ac{UAV} defines its own base control laws, state spaces, actuator mappings, and physical dynamics interactions.
Users can further store arbitrary parameters in a \ac{YAML} file for easy configuration.
The use of Python classes allows for a high degree of flexibility in defining complex interactions between the \acp{UAV} and their environment.

\subsection{Frame Conventions} \label{frame_conventions}
\begin{figure}
    \centering
    \includegraphics[width=0.24\textwidth]{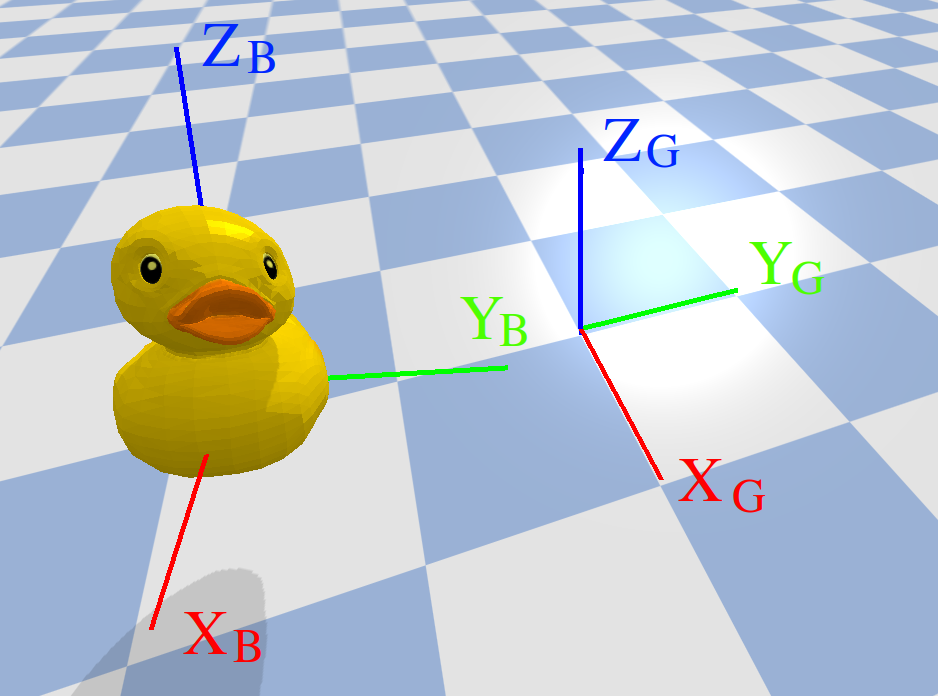}
    \caption{An illustration of body (subscript $B$) and ground (subscript $G$) frame axes used within PyFlyt.}
    \label{duck_frame}
\end{figure}

Frame conventions are essential for describing the orientation and movement of objects in space.
We use two reference frames, the ground frame and the body frame, shown in Figure \ref{duck_frame}.
The ground frame defines three axes relative to the local horizontal plane and the direction of gravity.
The body frame defines the axes relative to the body of the \ac{UAV} being controlled \cite{phang2014systematic}.
We utilize the \ac{ENU} frame convention, wherein for the ground frame, the X-axis points East, Y-axis points North, and Z-axis points Up (East and North are just cardinal references).
On the body frame, the \ac{ENU} convention defines the X, Y, and Z axes to point out the front, left, and upward direction from the \ac{UAV} respectively.

To transform a vector $\mathbf{v}_G \in \mathbb{R}^3$ from the ground frame to the body frame $\mathbf{v}_B \in \mathbb{R}^3$, we can use the following equation:
\begin{equation}
\mathbf{v}_\text{XYZ} = \mathbf{R} \times \mathbf{v}_\text{ENU}
\end{equation}
Where $\mathbf{R}$ denotes the rotation matrix from ground to body frame:
\begin{equation}
\mathbf{R} =
\begin{bmatrix}
c_\theta c_\psi && s_\phi c_\theta && -s_\theta \\
-s_\psi c_\phi + c_\psi s_\theta s_\phi && c_\psi c_\phi + s_\psi s_\theta s_\phi && c_\theta s_\phi \\
s_\psi s_\phi + c_\psi s_\theta c_\phi && -c_\psi s_\phi + s_\psi s_\theta c_\phi && c_\theta c_\phi \\
\end{bmatrix}
\end{equation}
Where $\phi$, $\theta$, and $\psi$ are the Euler angles describing the rotation between the two frames, $c_*$ and $s_*$ are shorts for $\cos{(*)}$ and $\sin{(*)}$ respectively.

\subsection{UAV Component Descriptions} \label{uav_component_descriptions}

The field of engineering often shares common mathematical models across similar components.
For example, the performance difference between two DC electric motors with varying stator heights are typically limited to parameters such as motor ramp time constant, maximum rotational speed, and torque output \cite{hemati1992complete}.
To provide a clear abstraction, PyFlyt offers configurable implementations of common components used in \ac{UAV} construction.
These components include vectorized implementations for brushless DC motor driven propellers, electric gimbals, fueled boosters, plus non-vectorized implementations for flapped lifting surfaces and cameras.
In this section, we describe the modelling of each component and their configurations.

\subsubsection{Brushless DC Motor Driven Propellers}

A brushless DC motor driven propeller, simply termed motor here, is a propulsion unit consisting of a fixed pitch propeller paired to a brushless DC electric motor, commonly seen on various \acp{UAV} as a means of propulsion.
We model the motor's RPM $\omega$ as a function of the throttle input $\zeta \in [-1, 1]$, using a first-order transfer function plus noise depending on the motor time constant $\tau_m$, maximum RPM $k_\omega$ and motor fluctuation standard deviation $\sigma_m$:
\begin{equation}
    \omega_{t+1} = \left( \frac{1}{\tau_m r_p} (\zeta k_\omega - \omega_t) + \omega_t \right)  (1 + \sigma_m \omega_t \eta)
\end{equation}
Where the subscript $t$ denotes the physics timestep, $r_p$ denotes the physics looprate in Hz, and $\eta$ is noise sampled from a standard Normal.
Depending on the thrust coefficient $k_F$ and torque coefficient $k_T$, this gets turned into a torque and force on the motor:
\begin{equation}
    T_m = k_T \omega ^ 2
\end{equation}
\begin{equation}
    F_m = k_F \omega ^ 2
\end{equation}

\subsubsection{Fueled Boosters}

Unlike electric motors, fueled boosters are propulsion units that produce no meaningful torque around their thrust axis and have limited throttleability.
More crucially, they depend on a fuel source that depletes with usage, changing the mass and inertia properties of the \ac{UAV}.
Additionally, some boosters, typically of the solid fuel variety \cite{abdelraouf2022thrust}, cannot be extinguished and reignited, a property we call reignitability.

In PyFlyt, the booster is parameterized with total fuel mass $m_{b_\text{max}}$, maximum fuel mass flow rate $\dot{m}_{b_\text{max}}$, the diagonal elements of the moment of inertia matrix at full fuel load  $J_{xx_\text{max}}$, $J_{yy_\text{max}}$, $J_{zz_\text{max}}$, minimum thrust $T_{b_\text{min}}$, maximum thrust $T_{b_\text{max}}$, a boolean for reignitability $\mathbb{I}_b$, the booster ramp time constant $\tau_b$ and a thrust output fluctuation standard deviation $\sigma_b$.
The booster component also allows placement of the fuel tank at an arbitrary location on the \ac{UAV}, useful for simulating jet fuel powered turbojet \acp{UAV} which have fuel stored in the main body and wings.

Fueled boosters have complex mathematical models that drastically change depending on altitude, temperature, total fuel remaining, and throttle setting \cite{d2001ignition, barato2015numerical, greatrix1995acceleration}.
In favour of simplicity, we adopt a simpler model for the booster in PyFlyt, useful as a baseline for developing more complex boosters should the user see necessary.
The control inputs to the booster include a boolean ignition state $\iota$ and a normalized throttle setting $\beta \in [0, 1]$.
The minimum duty cycle is defined as the ratio of minimum thrust to maximum thrust:
\begin{equation}
    \lambda_\text{min} = \frac{k_{t_\text{min}}}{k_{t_\text{max}}}
\end{equation}
Then, the current duty cycle of the booster $\lambda \in [0, 1]$ changes according to the throttle setting through a first order transfer function, plus some noise modelled similarly to that of the motor:
\begin{equation}
    \lambda_{t+1} =
    \mathbb{I}_{\lambda} \mathbb{I}_{m_b > 0}
    \left(
    \frac{1}{\tau_b r_p} (\beta (1 - \lambda_\text{min}) - \lambda_t) + \lambda_t
    \right)
    (1 + \sigma_b \lambda_t \eta) \numberthis \label{placeholder}
\end{equation}
Where $\mathbb{I}_{\lambda}$ is an indicator on whether the booster is lit while $\mathbb{I}_{m_b > 0}$ indicates the availability of remaining fuel.

The fuel consumption $\dot{m}_b$ of the booster depends on the duty cycle:
\begin{equation}
    \dot{m}_b = -\lambda k_{fr}  \mathbb{I}_{\lambda} \mathbb{I}_{m_b > 0}
\end{equation}

To prevent reignition of un-reignitable boosters, the booster lit indicator is recomputed at every timestep:
\begin{equation}
    \mathbb{I}_{\lambda, t+1} = (\neg \mathbb{I}_b \land \mathbb{I}_{\lambda, t}) \lor \iota_t
\end{equation}
In short, a booster that is not reignitable cannot be extinguished, while one that is reignitable can be extinguished and reignited at will.

From here, the thrust output depends on the duty cycle:
\begin{equation}
    T_b = \lambda T_{b_\text{max}}
\end{equation}
And inertia properties of the fuel tank depend on the remaining fuel balance:
\begin{equation}
    J_i = \frac{m_b}{m_{b_\text{max}}} J_{i_\text{max}} ; i \in \{xx, yy, zz\}
\end{equation}

There are several deficiencies with this model.
First, it does not model the sloshing of liquid fuel within the fuel tank.
Second, it does not model engine deterioration with the number of reignitions.
Third, solid rocket fuel boosters can have non-linear thrust profiles depending on the amount of fuel remaining \cite{puskulcu20083}, this is not modelled in the default booster model.
Lastly, there is presently no limit on the number of times a booster can be reignited, unlike most liquid-fueled rocket engines \cite{ma2018trajectory}.
That said, it is very difficult to have a parameterised booster model that suits the wide range of possible boosters in literature.
For that reason, the booster component is designed with generalisability and simplicity in mind; interested users are encouraged to modify the model depending on their needs.

\subsubsection{Electric Gimbals}

An electric gimbal is a powered joint that rotates around two axes.
The first servo motor rotates the joint about the first axis, $\mathbf{c}_a$, referenced to the body frame.
Similarly, the second servo motor rotates the joint about the second axis, $\mathbf{c}_b$, also referenced to the body frame.
Typically, the two rotational axes are orthogonal to each other, although this is not a necessity.
There are several uses for such gimballing, but the most common application is for thrust vectoring of propulsion systems \cite{cen2015integrated, kumar2020quaternion, papachristos2014efficient}, especially in scaled model rocketry \cite{jain2020design, kamath2021robust}.

The gimbal is controlled using a normalized input $\xi_i \in [-1, 1]; i \in {a, b}$ and prpoduces three dimensional rotation matrix about the \ac{UAV}'s body frame.
Configurable parameters include the unit vectors for the two rotational axes $\mathbf{c}_a, \mathbf{c}_b$, maximumum gimballing angles $k_{g-a}, k_{g-b}$, and actuation time constant $\tau_g$.
We utilize the same first order transfer function when converting input setpoint to actual gimbal angles $\mathbf{\epsilon}$:
\begin{equation}
    \epsilon_{i_{t+1}} = \frac{1}{\tau_g r_p} (\xi_i k_{g-i} - \epsilon_{i_t}) + \epsilon_{i_t}
\end{equation}
Where $i \in \{a, b\}$ is an axis index.
To obtain the final rotation matrix, Rodrigues' rotation formula is used \cite{valdenebro2016visualizing}.
Let the unit vector for an axis of rotation be $\mathbf{c}_i = \langle u_x, u_y, u_z \rangle$.
We then define $\mathbf{W}$ as:
\begin{equation}
    \mathbf{W}_i =
    \begin{bmatrix}
        0 & -u_z & u_y \\
        u_z & 0 & -u_x \\
        -u_y & u_x & 0
    \end{bmatrix}
\end{equation}
The Rodrigues' rotation matrix is then constructed as:
\begin{equation}
    \mathbf{R}_i = \mathbf{I} + \sin{\epsilon_i} \mathbf{W}_i + \left(2 \sin^2{\frac{\epsilon_i}{2}}\right) \mathbf{W}_i^2
\end{equation}
Where $\mathbf{I}$ is the identity matrix.
The final rotation matrix is thus:
\begin{equation}
    \mathbf{R} = \mathbf{R}_a \times \mathbf{R}_b
\end{equation}

The term multiplying $\mathbf{W}_i^2$ is typically written as $1 - \cos (\epsilon_i)$.
However, the sine variant is more numerically stable for values near $2n\pi; n \in \mathbb{Z}$.
Since $\mathbf{W}_i$ and $\mathbf{W}_i^2$ only depend on the axis of rotation which is predetermined at initialization, they can be precomputed and stored, resulting in a very fast computation for the rotation matrix around each axis.

\subsubsection{Flapped Lifting Surface}

Lifting surfaces are effectively flat plates with an aerofoil cross section that travel through the air.
In doing so, they experience a net force resulting in a lifting force $F_{s_\text{lift}}$, a drag force $F_{s_\text{drag}}$ and a net moment $T_s$ about the aerodynamic center.

The forces and moments that act on them depend on a number of factors, most notably the angle-of-attack $\alpha$ and freestream velocity $V_{\infty}$ according to the equation:
\begin{equation}
    \begin{bmatrix}
        F_{s_\text{lift}} \\
        F_{s_\text{drag}} \\
        T_s
    \end{bmatrix}
    =
    \begin{bmatrix}
        C_L \\
        C_D \\
        C_M \\
    \end{bmatrix}
    \frac{1}{2} \rho A_s V_{\infty}^2
\end{equation}
Where $A_s$ is the area of the lifting surface and $C_L$, $C_D$ and $C_M$ represent dimensionless lift, drag, and pitching moment coefficients, forming what we term here as a coefficient vector.

The coefficient vector can be captured in a lift, drag, and polar curve --- graphical representations of how the values vary with \ac{AoA}.
These curves, also known as a performance profile, are unique to each aerofoil and are often obtained through extensive \ac{CFD} simulation or wind tunnel testing.

In our approach, we adopt the method proposed by \cite{7152411} to obtain the performance profile.
Specifically, a lifting surface can be parameterized by its zero-lift \ac{AoA} $\alpha_0$, positive stall \ac{AoA} $\alpha_{\text{stall}_+}$, negative stall \ac{AoA} $\alpha_{\text{stall}_-}$, zero-lift drag coefficient $C_{D_0}$, lift coefficient slope $C_{L_\alpha}$, and a dimensionless viscosity correction factor $k_{s_\eta}$ \footnote{Page 112 of \cite{mccormick1994aerodynamics}.}.
The exact equations for curve generation are quite involved, interested readers are instead directed to the exact equations in \cite{7152411} or to implmentations in code \footnote{\url{https://github.com/jjshoots/PyFlyt/blob/master/PyFlyt/core/abstractions/lifting_surfaces.py\#L74}}.

The lifting surfaces in PyFlyt can have actuated flaps, hence the term flapped lifting surface.
This flap is parameterized with a flap-to-chord ratio --- a ratio of how much of the lifting surface is movable, a flap deflection limit $\delta_\text{max}$, and an actuation time constant $\tau_s$.
The flap allows the lifting surface to change its aerofoil geometry, consequently altering its performance profile curves.
We employ an approximation by \cite{khan2015real} to model this modification and utilize the familiar first order transfer function to model the flap deflection $\delta$ as a function of time and normalized actuation input $\nu \in [-1, 1]$:
\begin{equation}
    \delta_{t+1} =
    \frac{1}{\tau_s r_p}
    (\nu \delta_\text{max} - \delta_t) + \delta_t
\end{equation}

Our approximation is generally valid for normal operating conditions, but may not accurately model the aerodynamic behavior of the lifting surface in extreme conditions, such as when experiencing major side slip, highly turbulent air, or supersonic flows.

\subsubsection{Camera}

The camera component allows the development of \acp{UAV} with vision-based capabilities.
To further expand on this idea, there are no limits to the number of cameras a \ac{UAV} can have, and no restrictions on their locations relative to the \ac{UAV}.
This flexibility allows simulating stereo vision for depth perception \cite{iyengar2021simulator}, multiple viewpoints for collision avoidance \cite{akinola2021clamgen}, or multi-camera resolution \cite{lee2022reference}.

The camera in PyFlyt is parameterized with an image pixel height $h$ and width $w$, field of view in degrees, camera orientation and pose relative to the \ac{UAV}'s body frame, and whether the camera is gimbal stabilized.
The option for gimbal stabilization forces the camera to utilize rotation about the ground frame X and Y axes, providing a locked horizon view with the viewpoint following the body frame yaw.

When queried, each camera returns a set of 3 images readily obtained from the Bullet Physics Engine --- an RGB image in $\mathbb{R}^{w \times h \times 4}$, a depth map in $\mathbb{R}_+^{w \times h \times 1}$, and a segmentation map in $\mathbb{Z}^{w \times h \times 1}$ with values indicating the ID of the entity from which the pixel came from.

\subsection{UAV Model Description} \label{uav_model_description}

Several default \ac{UAV} models built using the components discussed in Section \ref{uav_component_descriptions} are shipped with PyFlyt.
Furthermore, it is possible to spawn multiples of each \ac{UAV} in the same environment and control them simultaneously with dynamic interactions.
In this section, we detail the design of three particular models that come with PyFlyt - a quadrotor, a fixedwing, and a rocket.

\subsubsection{QuadX}

The \textit{QuadX} \ac{UAV} describes a multirotor \ac{UAV} in the Quad-X configuration as described by ArduPilot \footnote{\url{https://ardupilot.org/copter/docs/connect-escs-and-motors.html\#quadcopter}} and PX4 \footnote{\url{https://dev.px4.io/v1.10/en/airframes/airframe_reference.html\#quadrotor-x}}.
In general, all quadrotor \acp{UAV} share a common mathematical model, albeit with varying physical and geometric parameters, such as mass, inertia, and size, as well as corresponding control parameters.
PyFlyt provides a generic implementation of the QuadX model utilizing the vectorized motor component implementation.
To construct arbitrary QuadX \acp{UAV}, a \ac{URDF} and \ac{YAML} file pair is simply passed to the underlying QuadX class to implement all mass, geometric, and control parameters of a QuadX \ac{UAV}.
As a result, users are able to construct QuadX \acp{UAV} without needing to modify the underlying Python code.

In addition, the PyFlyt implementation of the QuadX \ac{UAV} comes with various default control modes that researchers can build off  using the native \texttt{register\_controller} method.
These control modes utilize a cascaded \ac{PID} architecture \cite{phang2012design} to achieve increasingly higher levels of attitude control.
The implemented controllers allows for the following setpoint definitions:
\begin{itemize}
    \item Raw motor outputs
    \item Angular velocity control and normalized thrust output
    \item Angular position control and climb rate
    \item Linear velocity control and yaw rate
    \item Linear velocity control and yaw angle
    \item Position control and yaw rate
    \item Position control and yaw angle
\end{itemize}

As a reference, we provide two different QuadX models in PyFlyt.
The first is the model of a Bitcraze Crazyflie 2.x, leveraging the extensive amounts of system identification work available \cite{landry2015planning, luis2016design, forster2015system}.
However, instead of adhering to a true 1:1 representation of the Crazyflie quadrotor, we modify the model to have an 8:1 thrust to weight ratio, allowing for more interesting and dynamic performance characteristics in downstream \ac{RL} tasks.
The second model is that of a generic 1 kg quadrotor \ac{UAV} with parameters mimicking a generic F450 quadrotor \ac{UAV} \footnote{\url{https://www-v1.dji.com/flame-wheel-arf/feature.html}}, meant to serve as a starting point for users implementing models of their own without using a \ac{DAE} file.

\subsubsection{Fixedwing}

The \textit{Fixedwing} \ac{UAV} is a tube-and-wing aircraft consisting of five lifting surface components---main wing, left aileron, right aileron, horizontal tail, vertical tail---and a motor.
All of the lifting surfaces except the main wing are flapped, and they have parameters of an ideal 2D aerofoil.
The ailerons have a maximum flap deflection of 30$^{\circ}$, while the rudder and elevators are limited to 20$^{\circ}$.
It has a wingspan of 2.4 m, a nose to tail length of 1.55 m, with a 0.8 thrust to weight ratio.
The total mass of the \ac{UAV} is 2.35 kg, with 1 kg uniformly distributed in the fuselage, 0.5 kg in the main wing and the remaining mass distributed among the other lifting surface components.

\subsubsection{Rocket}

The \textit{Rocket} \ac{UAV} is  a 1:10th scale rocket closely modeled after the design of SpaceX's Falcon 9 \footnote{\url{https://www.spacex.com/vehicles/falcon-9/}} v1.2 first stage and interstage.
Mass and geometry properties were extracted from a datasheet by Space Launch Report \footnote{\url{https://web.archive.org/web/20170825204357/spacelaunchreport.com/falcon9ft.html\#f9stglog}}.
It consists of booster, gimbal, and lifting surface components.
The SpaceX Falcon 9 features grid fins \cite{washington1993grid} as supplementary control actuators.
However, at the time of writing, limited studies on the aerodynamic properties of the Falcon 9's grid fins were made available to the public \cite{lee2020aerodynamic}.
As a consequence to this shortcoming, the Rocket's control fin properties were simply approximated.
In addition, the Bullet Physics Engine approximating the world as a flat 2D surface, it is not feasible to simulate the orbital mechanics of the Rocket within the PyFlyt framework.

\subsection{Gymnasium Environments} \label{gymnasium_environments}

The primary goal of PyFlyt is to facilitate \ac{RL} research for \acp{UAV}.
To that end, we provide several default \ac{RL} environments utilizing the Gymnasium API for a variety of \ac{UAV} models with various tasks and reward functions.
This section briefly covers the mathematical model of a \ac{MDP} and the Gymnasium API, before providing details about the various environments provided.

\subsubsection{Markov Decision Processes}

\begin{figure}
    \centering
    \includegraphics[width=0.3\textwidth]{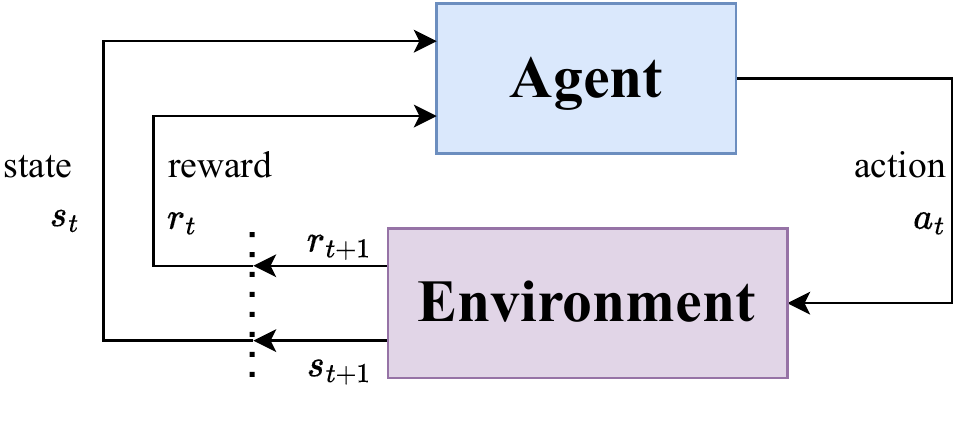}
    \caption{A classic model of an \ac{MDP} represented as a cyclic graph.}
    \label{mdp}
\end{figure}

The Markov Decision Process (MDP) is a mathematical framework used to model decision-making in situations where outcomes are uncertain.
It consists of a set of states $\mathbf{s} \in \mathbf{S}$, actions $\mathbf{a} \in \mathbf{A}$, and rewards $r \in \mathbb{R}$.
At each time step, the agent chooses an action based on the current state, which leads to a new state and a reward.
The goal is to find a policy that maximizes the expected cumulative reward over time.
Figure \ref{mdp} illustrates the \ac{MDP} as a cyclic graph.

\subsubsection{Gymnasium API}

Gymnasium is a standard API for defining \ac{RL} environments.
It is the maintained fork of the OpenAI Gym library, with a simple Pythonic interface.
The Gymnasium API allows for the universal definition of state and action spaces through a set of default constructs.
It also defines environment termination and truncation behaviour, convenient environment wrappers, as well as a plethora of native environments from which users can use as benchmarks or use as reference.

Generally, libraries that utilize the Gymnasium API can be segregated into one of two parties --- learning libraries which implement a suite of well-studied and effective learning algorithms, and environment libraries which allow for the benchmarking and construction of \ac{RL} problems.
The existence of such a standardized API allows for the application of general purpose \ac{RL} algorithms to arbitrary problems.
In this context, PyFlyt is a means of standardizing the construction of \acp{UAV} for \ac{RL} problems, with which we provide several default environments built to match the Gymnasium API.

\subsubsection{Base Environment Parameters}

\paragraph{\textbf{Observation Spaces}}

All currently implemented environments within PyFlyt utilize either the \texttt{Box} space, or a \texttt{Dict} space consisting of a \texttt{Box} space and a \texttt{Sequence} space.
The \texttt{Box} space defines a space in $\mathbb{R}^{n \times ...}$, while a \texttt{Sequence} space defines a space of a sequence of elements in $\mathbb{R}^{n \times ...}$.
Meanwhile, the \texttt{Dict} space is a means of combining various spaces together to form a composite space.

In Pyflyt, the \ac{UAV}'s attitude, previous action, and component statuses are represented as a \texttt{Box} space.
Waypoint reaching environments have targets defined as a \texttt{Sequence} space, which is combined with the aforementioned \texttt{Box} state through a \texttt{Dict}.
More concisely, environments without waypoints have the following state space:
\begin{equation}
    \mathbf{s}_\text{attitude} = [ \dot{\mathbf{\theta}}_B, \mathbf{\theta}_G, \dot{\mathbf{p}}_B, \mathbf{p}_G, \mathbf{a}_{t-1}, \mathbf{s}_\text{aux} ]
\end{equation}
Where $\theta$ represents the angular position of the \ac{UAV} and $p$ represents the linear position, both referenced to either body (subscript $B$) or ground (subscript $G$) frames.
In addition, $\mathbf{s}_\text{aux}$ refers to auxiliary information, such as normalized lifting surface deflection angle, motor duty cycle, or booster fuel and ignition statuses.

For environments with waypoints, the state space has the following form:
\begin{equation}
    \mathbf{s} = \{ \mathbf{s}_\text{attitude}, \mathbf{s}_\text{target\_deltas} \}
\end{equation}
Where $\mathbf{s}_\text{target\_deltas}$ represent the sequence of waypoints remaining in the environment:
\begin{equation}
    \mathbf{s}_\text{target\_deltas} = [ \Delta \mathbf{p}_1^T, \Delta \mathbf{p}_2^T, ..., \Delta \mathbf{p}_n^T ]
\end{equation}
Where $\Delta \mathbf{p}_i$ represent the vector to the $i$-th remaining waypoint defined in the \ac{UAV}'s body frame.

\paragraph{\textbf{Action Spaces}}

All action spaces for currently implemented environments use a \texttt{Box} action space.
For QuadX environments, the action is a four wide vector representing roll, pitch and yaw rates along with a normalized thrust command.
For Fixedwing environments, the action is a four wide vector representing normalized aileron, horizontal and vertical stabiliser deflection angles along with a normalized thrust command.
For Rocket environments, the action is a seven wide vector representing grid fin deflections in the X plane, Y plane, and cyclical deflections, along with the booster ignition, throttle settings and two values for booster gimballing about the body frame XY axes.

\paragraph{\textbf{Looprates}} All environments operate with a physics looprate of 240 Hz while controllers run at 120 Hz---allowing UAV models to operate lower level controllers under the hood.
At the environment level, all interactions between an agent and the environment happen at a reconfigurable rate of 30 Hz.
This configuration allows the \ac{RL} agents to send setpoints to the \acp{UAV} instead of raw actuator commands, a scheme similar to human pilots for racing quadrotor \acp{UAV} \cite{tai2018application}.

\subsubsection{Reward Functions}

In \ac{RL}, the reward function directly influences the learning dynamics and final policy obtained from a particular task.
A dense reward function is one that provides consistent feedback to the agent about its actions.
This means that the reward signal is provided at almost every time step, and the agent can easily understand the impact of its actions on the environment.

On the other hand, a sparse reward is one that is given infrequently and provides limited information to the agent about its actions.
This means that the reward signal is only provided occasionally or at certain milestones, making it harder for the agent to understand the impact of its actions on the environment.
Sparse rewards are commonly encountered in real-world tasks where the reward function is not immediately obvious and may require additional exploration in different parts of the environment to discover it.

In general, dense rewards can encourage an agent to learn quickly and converge to an optimal solution.
However, sparse rewards often lead to a more robust and generalised solution as the agent is forced to learn more about the environment through additional exploration.
In addition, sparse rewards represent a more holistic view of the goal of the underlying task, leading to agents that are more focused on achieving the primary objective without getting distracted by irrelevant factors \cite{eschmann2021reward}.
All of the default environments in PyFlyt come with a dense and sparse reward variants, available through the \texttt{sparse\_reward} argument on environment initialization.

\subsubsection{Provided Environments}

At the time of writing, there are four pre-built environments in PyFlyt, described in this section.

\paragraph{\texttt{PyFlyt/QuadX-Hover-v0}}

The task of this environment is for an agent to hover the QuadX \ac{UAV} indefinitely.
In the dense setting, the reward of the environment is described as follows:
\begin{equation}
    r =
    \begin{cases}
        -100 & \text{if crash}, \\
        -||[\theta, \phi]|| - ||[x, y, z-1]|| & \text{otherwise}
    \end{cases}
\end{equation}
Where $\theta$ and $\phi$ are the pitch and roll angles in radians, $x$, $y$ and $z$ describe the linear position of the \ac{UAV}.

In the sparse reward variant, the reward is simply -0.1 for each timestep and -100 for crashing.

\paragraph{\texttt{PyFlyt/QuadX-Waypoints-v0}}

This environment describes the task of getting a QuadX \ac{UAV} to reach a series of randomly generated waypoints in 3D space in a specific order.
By default, there are five waypoints in total, but this number is reconfigurable.
The dense reward function of the environment is as follows:
\begin{equation}
    r =
    \begin{cases}
        -100 & \text{if crash}, \\
        c_a \delta^{-1} - c_b \dot{\delta} & \text{if } \dot{\delta} < 0, \\
        100 & \text{if waypoint reached}, \\
        \delta^{-1} & \text{otherwise}
    \end{cases}
\end{equation}
Where $\delta$ is the displacement from the \ac{UAV} to the next target, and $c_a$ and $c_b$ are scaling coefficients, set at 0.1 and 3 respectively.
To avoid reward function blowup caused by the $\delta^{-1}$ component, a waypoint is considered reached when the \ac{UAV}'s center of mass is some distance $d$ away from the waypoint.
This distance is a reconfigurable parameter that is set at 0.2 m by default.
In the sparse reward setting, the reward is simply -100 for crashing and 100 for reaching a waypoint.

\paragraph{\texttt{PyFlyt/Fixedwing-Waypoints-v0}}

This environment is similar to \texttt{PyFlyt/QuadX-Waypoints-v0} but is modified to use the Fixedwing \ac{UAV}.
Due to the increase in size of the \ac{UAV} model, $d$ is set at 2, $c_a$ is set at 1 and $c_b$ is set at 3.
In addition, the waypoints are also generated much further apart, at a scale of 20 times that of the QuadX environment.

\paragraph{\texttt{PyFlyt/Rocket-Landing-v0}}

The objective in this environment is to safely land the Rocket \ac{UAV} model on a 4-meter diameter pad from an initial drop height of 450 meters---about 100$\times$ the height of the Rocket. 
The landing is considered successful if the rocket contacts the pad with a velocity of less than 1 m/s and comes to an upright halt.

In the dense reward setting, the rocket is penalized according to the absolute distance between the center of the rocket and the landing pad, $d_\text{pad}$:
\begin{equation}
    r_d = -0.2 \cdot d_\text{pad}
\end{equation}
Note that $d_\text{pad} > 0$ as it is the distance from the center of mass of the rocket to the landing pad.
We further define a penalty for having the rocket favour lower absolute velocities when near the landing pad:

\begin{equation}
    r_v = -d^{-2}_\text{pad} \cdot v
\end{equation}
Finally, a termination reward was computed based on the landing condition:
\begin{equation}
    r_\text{pad} =
    \begin{cases}
        -20 & \text{if fatal}, \\
        100 & \text{if safe}
    \end{cases}
\end{equation}
Combined, this leads to a reward function of:
\begin{equation}
    r = r_d + r_v + \mathbb{I}_\text{pad} r_\text{pad} - 100 \mathbb{I}_\text{ground}
\end{equation}
Where $\mathbb{I}_\text{pad}$ indicates whether the rocket is on the landing pad and $\mathbb{I}_\text{pad}$ is an indicator for ground contact.
More concisely, we penalize the rocket heavily for landing outside of the landing pad, while awarding policies that favour low velocities near the landing pad leading to safe touchdowns.
To increase environment difficulty and prevent the execution of otherwise irrelevant manoeuvres, the Rocket starts with only 1\% of total fuel---just enough for one landing attempt.

\section{Reinforcement Learning Results} \label{reinforcement_learning_results}

\begin{figure}
    \centering
    \includegraphics[width=0.2\textwidth]{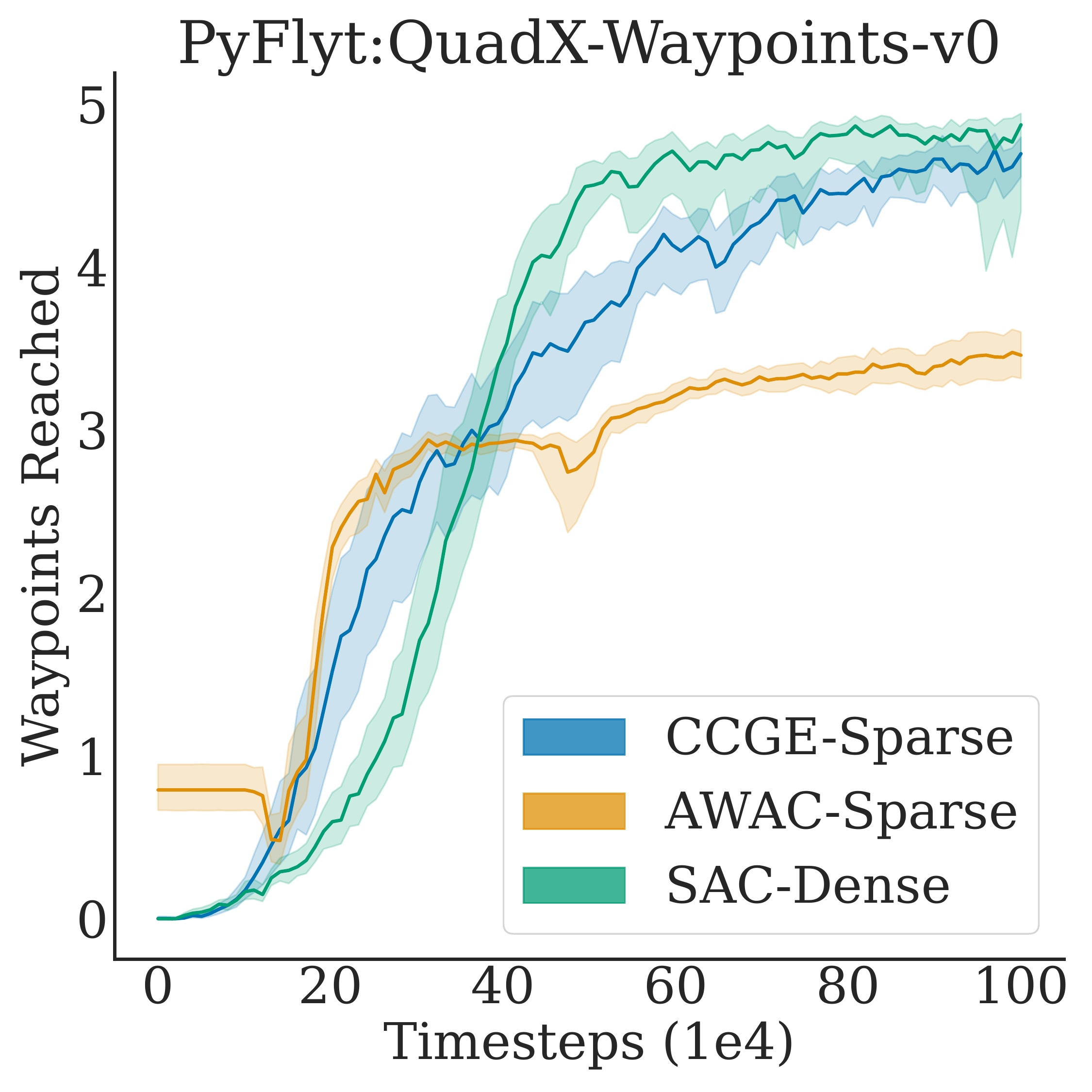}
    \includegraphics[width=0.2\textwidth]{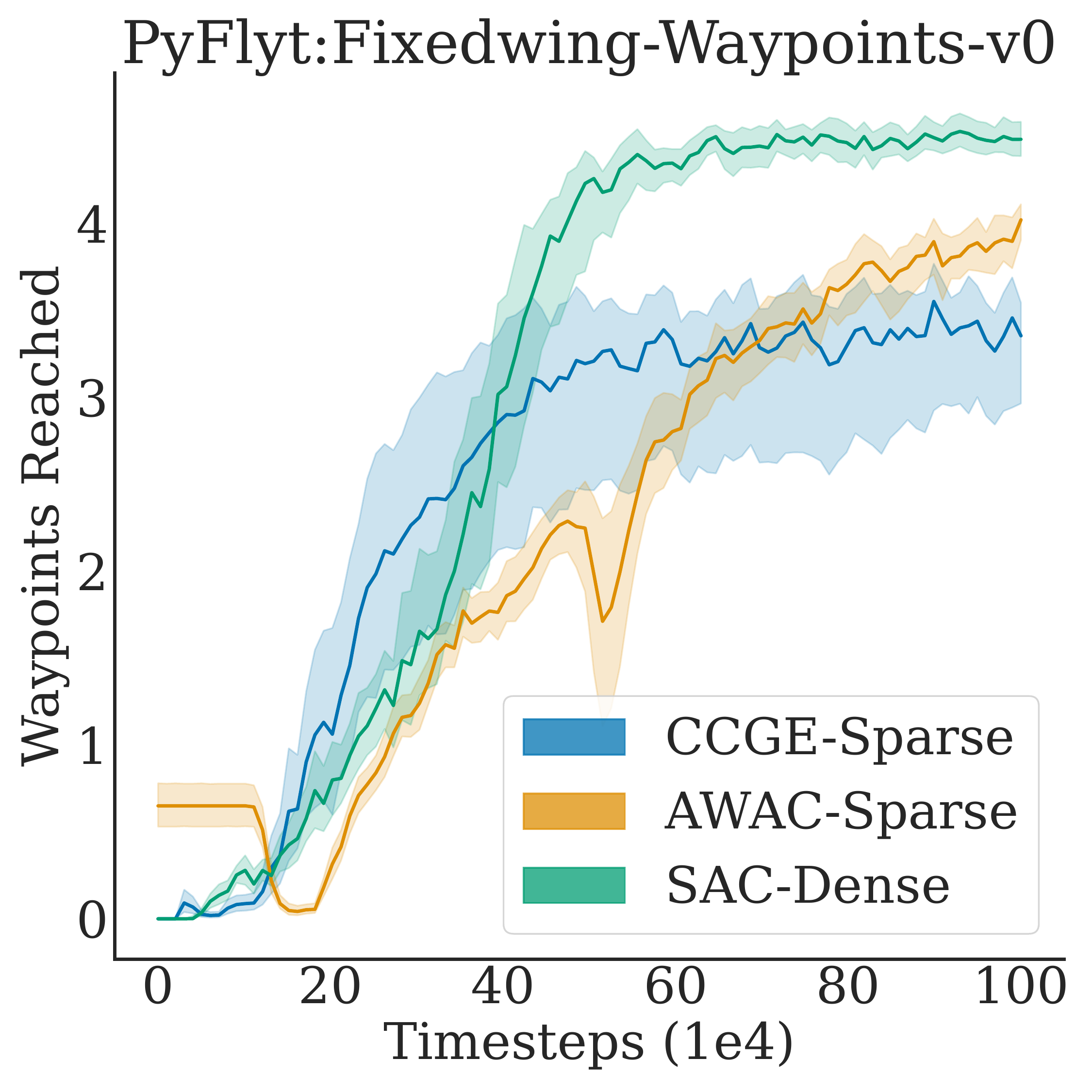}
    \caption{Results of training \ac{RL} agents in PyFlyt on the different environments.}
    \label{rl_results}
\end{figure}

\begin{figure}
    \centering
    \includegraphics[width=0.23\textwidth]{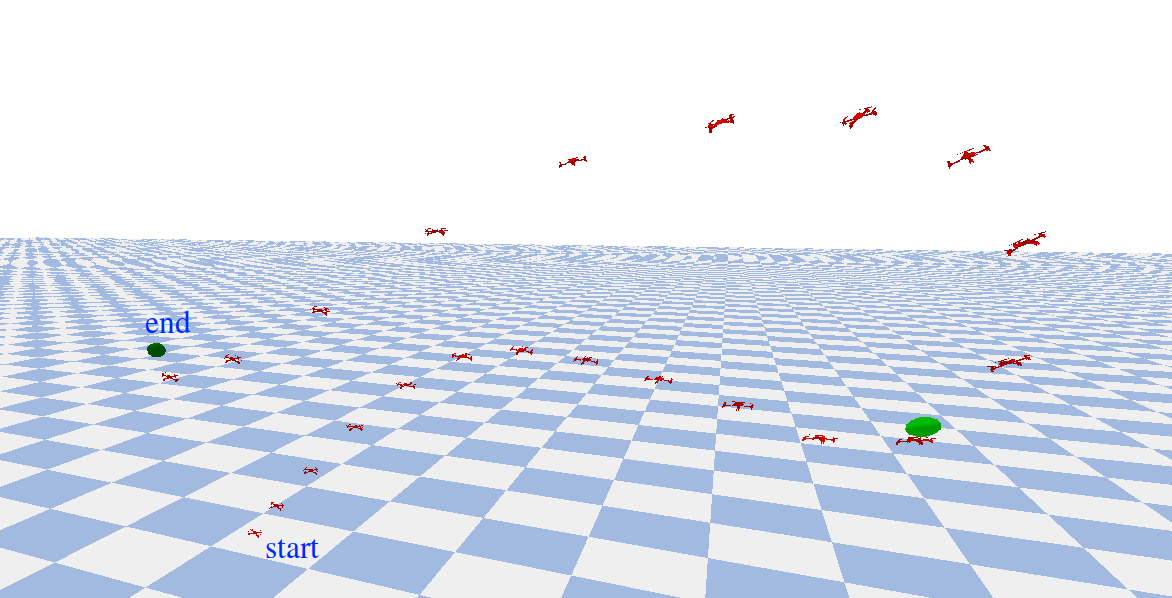}
    \includegraphics[width=0.23\textwidth]{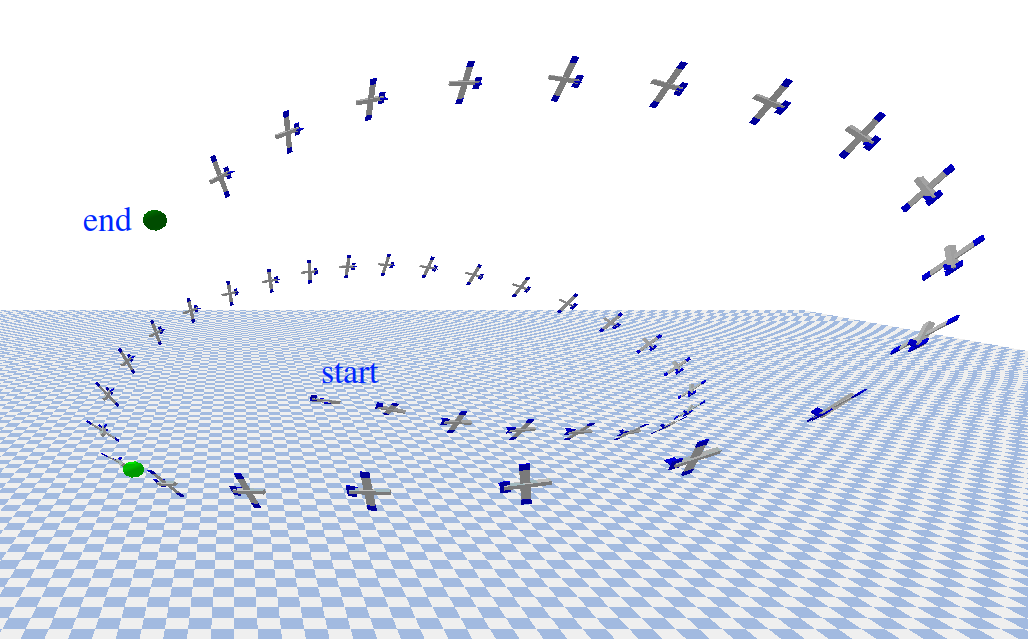}
    \caption{Example trajectories of trained agents in \texttt{PyFlyt/QuadX-Waypoints-v0} and \texttt{PyFlyt/Fixedwing-Waypoints-v0}, both with only two waypoints for easier legibility.}
    \label{rl_trajectories}
\end{figure}

Preliminary results for training \ac{RL} algorithms in \texttt{PyFlyt/QuadX-Waypoints-v0} and \texttt{PyFlyt/Fixedwing-Waypoints-v0} are provided here, as we think these represent the an interesting class of problems for \ac{UAV} flight.
The environments were tested using \ac{SAC} \cite{haarnoja2018soft} in the  dense reward setting, while the sparse reward variants were tested using \ac{CCGE} \cite{jet2022ccge} and \ac{AWAC} \cite{nair2020awac} by bootstrapping off a policy partially trained using \ac{SAC}.
We obtain 50 learning curves for each environment-algorithm combination over different random seeds.
We then plot the interquartile means of the learning curves along with their bootstrapped confidence intervals using the RLiable library \cite{agarwal2021deep}.
Our results, shown in Figure \ref{rl_results}, are meant to show the feasibility of training agents in these environments across multiple different algorithms.
Examples of flight trajectories for each environment are also shown in Figure \ref{rl_trajectories} \footnote{Videos of trajectories are available at \url{https://github.com/jjshoots/PyFlyt/blob/master/readme.md}}.

\section{Possible Extensions and Future Work} \label{possible_extensions}

PyFlyt v1.0 has been released and is ready for use.
That said, further improvements have been considered as part of PyFlyt's development roadmap.
First, we plan to implement default creation of multi-agent environments for multi-agent \ac{RL}, with PettingZoo support as our primary focus \cite{terry2021pettingzoo}.
Various additional \ac{UAV} types are also currently being modelled, such as quadplane, flying wing, and hexacopter \acp{UAV}.
We also aim to introduce a number of aerodynamics associated simulation attributes, such as Dryden turbulence modeling, down wash, and aerodynamic shielding.
Finally, user experience and documentation are vital components of any library, we therefore aim to provide clear, concise documentation, examples, and tutorials for PyFlyt, with the intention of reducing the barrier of entry for new users and increasing its adoption within the community.

\section{Conclusion} \label{conclusion}

This paper introduced PyFlyt, a novel open-source simulation tool designed for machine learning and robotics workflows in \ac{UAV} research.
With its Pythonic interface and modular implementation of \ac{UAV} components, PyFlyt enables the configuration of arbitrary \ac{UAV} types, and offers prebuilt \ac{UAV} implementations.
A number of default Gymnasium environments were also introduced with both dense and sparse reward variants.
We further present successful completion of the default environment by three distinct \ac{RL} algorithms within the PyFlyt environment, underscoring its potential as a valuable tool for future \ac{UAV} research.
We are confident that PyFlyt will facilitate innovative research in robotics and autonomous \ac{UAV} flight, and we look forward to see its utilization by the community.

\bibliographystyle{ieeetr}
\bibliography{references}


\section{Biography Section}

\begin{IEEEbiographynophoto}
{Jun Jet Tai} is a PhD student at Coventry University working in the field of reinforcement learning and unmanned aerial robotics.
Previously, he obtained his undergraduate's degree in Mechanical Engineering from Taylor's University.
While there, he worked as a research assistant under the Taylor's Unmanned Aerial Vehicles research group under Dr. Swee King Phang, where he designed novel control and obstacle avoidance algorithms for quadrotor \acp{UAV}.
\end{IEEEbiographynophoto}

\begin{IEEEbiographynophoto}
{Jim Wong} is a final year Aerospace Engineering student at Coventry University. 
His final year research project focuses on Path Planning and Path Following control algorithms for a fixed wing UAV, with emphasis on optimal control and flight dynamics modelling.
He is also interested in the practical applications of artificial intelligence and machine learning in his research in the development of autonomous unmanned aerial vehicles.
\end{IEEEbiographynophoto}

\begin{IEEEbiographynophoto}
{Mauro S. Innocente} is an Assistant Professor in Aerospace Engineering, a member of the Centre for Future Transport and Cities, and the founder of the Autonomous Vehicles \& Artificial Intelligence Laboratory at Coventry University. He received his engineering degree from the National University of the Northeast (2003), his MSc degree in numerical methods from the Polytechnic University of Catalonia (2007), and his PhD degree from Swansea University (2011). His research focuses on mathematical modelling, computational intelligence, swarm intelligence, swarm robotics, reinforcement learning, and optimisation within a comprehensive spectrum of applications.
\end{IEEEbiographynophoto}

\begin{IEEEbiographynophoto}
{Nadjim Horri} received his PhD degree from the University of Surrey in 2004. He is currently a lecturer in aerospace control at the University of Leicester. He worked as a research fellow at the Surrey Space Centre and as an assistant professor in aerospace engineering at Coventry University. His research interests include the state and parameter estimation and control of manned and unmanned aircraft and spacecraft, including fault tolerance and the handling of flight constraints and perturbations.
\end{IEEEbiographynophoto}

\begin{IEEEbiographynophoto}
{James Brusey} (M'13) received the B.Ap.Sc. and Ph.D. degrees from RMIT University in 1996 and 2003, respectively. He is currently a Professor of Computer Science and co-Director of the Centre for Computer Science and Mathematical Methods at Coventry University, leading on AI for Cyberphysical Systems. His research interests include Reinforcement Learning applied to real-world, human-centric control systems. 
\end{IEEEbiographynophoto}

\begin{IEEEbiographynophoto}
{Swee King Phang} (SM'22) received the B.Eng. and Ph.D. degrees from the National University of Singapore, Singapore, in 2010 and 2014. He is currently a Senior Lecturer of School of Engineering, and hub leader for Automation and Robotics Hub at Taylor's University, Malaysia. His research interests are mainly on unmanned aerial systems, including modeling and flight controller design, indoor navigation and trajectory optimization.
\end{IEEEbiographynophoto}

\vfill

\end{document}